\documentclass[sn-mathphys]{sn-jnl}

\usepackage[shortlabels]{enumitem}
\usepackage{soul}  

\jyear{2021}%

\theoremstyle{thmstyleone}%
\theoremstyle{thmstyletwo}%

\theoremstyle{thmstylethree}%

\begin{document}

\title[To Raise or Not To Raise: The Autonomous Learning Rate Question]{To Raise or Not To Raise: The Autonomous Learning Rate Question}


\author*[1,2]{\fnm{Xiaomeng} \sur{Dong}}\email{Xiaomeng.Dong@ge.com}

\author[1]{\fnm{Tao} \sur{Tan}}

\author[1]{\fnm{Michael} \sur{Potter}}

\author[1]{\fnm{Yun-Chan} \sur{Tsai}}

\author[1]{\fnm{Gaurav} \sur{Kumar}}

\author[1]{\fnm{V. Ratna} \sur{Saripalli}}

\author[2]{\fnm{Theodore} \sur{Trafalis}}

\affil[1]{\orgname{GE HealthCare},\country{USA}}

\affil[2]{\orgname{University of Oklahoma}, \country{USA}}


\abstract{There is a parameter ubiquitous throughout the deep learning world: learning rate. There is likewise a ubiquitous question: what should that learning rate be? The true answer to this question is often tedious and time consuming to obtain, and a great deal of arcane knowledge has accumulated in recent years over how to pick and modify learning rates to achieve optimal training performance. Moreover, the long hours spent carefully crafting the perfect learning rate can come to nothing the moment your network architecture, optimizer, dataset, or initial conditions change ever so slightly. But it need not be this way. We propose a new answer to the great learning rate question: the Autonomous Learning Rate Controller. Find it at \url{https://github.com/fastestimator/ARC/tree/v2.0}.}


\maketitle

\section{Introduction}\label{sec1}

Learning Rate (LR) is one of the most important hyperparameters in deep learning training, a parameter everyone interacts with for all tasks. In order to ensure model performance and convergence speed, LR needs to be carefully chosen. Overly large LRs will cause divergence whereas small LRs train slowly and may get trapped in a bad local minima. As training schemes have evolved over time they have begun to move away from a single static LR and into scheduled LRs, as can be seen in a variety of state-of-the-art AI applications \cite{sota-transformer,sota-yolov4,sota-cspnet,sota-roberta}. LR scheduling provides finer control of LRs by allowing different LRs to be used throughout the training. However, the extra flexibility comes at a cost: these schedules bring more parameters to tune. Given this tradeoff, there are broadly two ways of approaching LR scheduling within the AI community.

Experts with sufficient computational resources tend to hand-craft their own LR schedules, because a well-customized LR schedule can often lead to improvements over current state-of-the-art results. For example, entries in the Dawnbench \cite{dawnbench} are known for using carefully tuned LR schedules to achieve world-record convergence speeds. However, such LR schedules come with significant drawbacks. First, these LR schedules are often specifically tailored to an exact problem configuration (architecture, dataset, optimizer, etc.) such that they do not generalize to other tasks. Moreover, creating these schedules tends to require a good deal of intuition, heuristics, and manual observation of training trends. As a result, building a well-customized LR schedule often requires great expertise and significant computing resources.

In contrast, others favor existing task-independent LR schedules since they often provide decent performance gains with less tuning efforts. Some popular choices are cyclic cosine decay \cite{cosinedecay}, exponential decay, and warmup \cite{lrwarmup}. While these LR schedules can be used across different tasks, they are not specially optimized for any of them. As a result, these schedules do not guarantee performance improvements over a constant LR. On top of that, many of these schedules still require significant tuning to work well. For example, in cyclic cosine decay, parameters such as $l_{max}$, $l_{min}$, $T_0$, and $T_{multi}$ must all be tuned in order to function properly. Recently some methods have been proposed which either provide techniques to easily infer their parameters (\emph{e.g.} super-convergence \cite{superconvergence}), or are even completely parameter free (\emph{e.g.} stochastic line search \cite{vaswani2019painless,mahsereci2017probabilistic}). While promising, these techniques can prove finicky in practice, as we demonstrate in Section \ref{sec:experiments}.

Recent advancements in AutoML on architecture search \cite{nas,enas,darts} and update rule search \cite{updaterule} have proved that it is possible to create automated systems that perform equal or better than human experts in designing deep learning algorithms. These successes have inspired us to tackle the LR scheduling problem. We aim to create a system that learns how to change LR effectively. Several attempts have been previously made to dynamically learn to set LRs \cite{LARS,shu2020metalrschedulenet,daniel2016learning}. Unfortunately these have introduced step level dual optimization loops into the training procedure, something which can add significant complexity and training time to a problem. Moreover, this family of techniques has been shown to be vulnerable to short-horizon bias \cite{wu2018understanding}, sacrificing long-term performance for short-term gains.  We believe that both of these drawbacks can be overcome.

To that end we introduce ARC: an Autonomous LR Controller. It takes training signals as inputs and is able to intelligently adjust LRs in a real-time generalizable fashion. ARC overcomes the challenges faced by prior LR schedulers by encoding experiences over a variety of different training tasks, different time horizons, and by dynamically responding to each new training situation so that no manual parameter tuning is required.

ARC is also fully complementary to modern adaptive optimizers such as Adagrad \cite{adagrad} and Adam \cite{adam}. Adaptive optimizers compute updates using a combination of LR and `adaptive' gradients. When gradients have inconsistent directions across steps, the scale of the adaptive gradient is reduced. Conversely, multiple updates in the same direction result in gradient upscaling. This is sometimes referred to as adaptive LR even though the LR term has not actually been modified. Our method is gradient agnostic and instead leverages information from various training signals to directly modify the optimizer LR. This allows it to detect patterns which are invisible to adaptive optimizers. Thus the two can be used in tandem for even better results.

The key contributions of this work are:

\begin{enumerate}
\item An overall methodology for developing autonomous LR systems, including problem framing and dataset construction.

\item A comparison of ARC with popular LR schedules across multiple computer vision and language tasks.

\item An analysis of failure modes and unexpected behaviors from ARC, informing future directions for research.
\end{enumerate}

The rest of the paper is organized as follows: Section \ref{sec:CC} outlines constraints which guide further solution development, Section \ref{sec:methods} explains the ARC methodology, Section \ref{sec:experiments} experimentally compares the performance of ARC against other common LR scheduling methods, and Section \ref{sec:limits} discusses open problems and opportunities.

\section{Challenges and Constraints}
\label{sec:CC}

Before delving into our methodology, we will first highlight some of the key challenges in developing an autonomous LR controller. These constraints inform many of our subsequent design decisions.

\begin{enumerate}[label=\alph*)]
\item \label{challenges:subjective} \textbf{Subjectivity.} Determining the superiority of one model over another (each trained with a different LR) is fraught with subjectivity. There are many different ways to measure model performance (training loss, validation loss, accuracy, etc.) and they may often be in conflict with one another.

\item \label{challenges:cumulative} \textbf{Cumulativeness.} Associating the current model performance with an LR decision at any particular step is challenging, since the current performance is the result of the cumulative effect of all previous LRs used during training. This is also related to short-horizon bias \cite{wu2018understanding}, where long-term effects are easily overshadowed by short-term wins.

\item \label{challenges:random} \textbf{Randomness.} Randomness during training makes it difficult to compare two alternative LRs. Some common sources of randomness are dataset shuffling, data augmentation, and random network layers such as dropout. Any performance differences due to the choice of LR need to be large enough to overshadow these random effects.

\item \label{challenges:scale} \textbf{Scale.} Different deep learning tasks use different metrics to monitor training. The most task-independent of these are training loss, validation loss, and LR. Unfortunately, the magnitude of these values can still vary greatly between tasks. For example, categorical cross entropy for 1000-class classification is usually between 0 and 10, but a pixel-level cross entropy for segmentation can easily reach a scale of several thousand. Moreover, a reasonable LR for a given task can vary greatly, from 1e-6 up to 10 or more.

\item \label{challenges:footprint} \textbf{Footprint.} The purpose of having an automated LR controller is to achieve faster convergence and better results. Any solution must therefore have a small enough footprint that using it does not adversely impact training speed and memory consumption.

\end{enumerate}

\section{Methods}
\label{sec:methods}

\subsection{Framing LR Control as a Learning Problem}

We frame the development of ARC as a supervised learning problem: predicting the next LR given available training history. Due to challenge \ref{challenges:cumulative}, the model needs to observe the consequence of a specific LR for long enough to form a clear association between LR and performance. We therefore only modify the LR on an epoch timescale. This has a secondary benefit of dramatically reducing our computational overhead compared with competing methods.

Per challenges \ref{challenges:subjective} and \ref{challenges:scale}, as well as the desire to create a generalizable system, we cannot use any task-specific metrics. We also cannot rely on model parameters or gradient inspection since ARC would then become architecture dependent and would likely also fail constraint \ref{challenges:footprint}. We therefore leverage only the historical training loss, validation loss, and LR as input features.

Due to challenges \ref{challenges:random} and \ref{challenges:scale}, rather than generating a continuous prediction of what new LR values should be, we instead pose this as a 3-class classification problem. Given the input features, should the LR: increase ($LR * 1.618$), remain the same ($LR * 1.0$), or decrease ($LR * 0.618$)? We chose these specific values based on the following intuitions: Let $\alpha$ be your desired increase factor and $\beta$ be your desired decrease factor. Then we would like $\alpha\beta=1$ such that you can easily undo a decision if it later turns out to have been a mistake. We also would like $\alpha$ not to be too large, since a single large jump in LR could easily trigger model divergence. $\alpha$ should likewise not be too small, since then you might miss an opportunity to properly take advantage of smooth loss landscapes. What then is the largest `safe' value for $\alpha$? While we have no perfect way of knowing this in the general case, one reasonable heuristic is to combine previous LR values which are already known to be safe. For example, to get a new increased LR, you could sum together the two largest LRs which you've already tested. In other words: $\alpha^{i+1}LR_0=\alpha^{i}LR_0+\alpha^{i-1}LR_0$. Solving for these two constraints gives $\alpha\approx 1.618$ and $\beta \approx 0.618$.

\subsection{Generating the Dataset}
\label{subsec:dataset}

Having framed the problem, we now need to generate a dataset on which we can train ARC. Given that this specific learning problem is newly proposed, there are no existing datasets nor data curation workflows which we could leverage. We therefore build one ourselves from real deep learning training tasks. For each task we used the following procedure to generate data:

\begin{enumerate}
\item \label{step:train} Train $n$ epochs with $LR=r$, then save the current state as checkpoint $C$

\item \label{step:increase} Reload $C$, train for $n$ epochs with $LR=1.618*r$, then compute validation loss ($l_+$)

\item \label{step:constant} Reload $C$, train for $n$ epochs with $LR=1.0*r$, then compute validation loss ($l_1$)

\item \label{step:decrease} Reload $C$, train for $n$ epochs with $LR=0.618*r$, then compute validation loss ($l_-$)

\item \label{step:truth} Note the $LR$ which resulted in $min\left\{ l_+, l_1, l_- \right\}$

\item \label{step:repeat} Reload $C$, eliminate $max\left\{ l_+, l_1, l_- \right\}$ and its corresponding $LR$, replace $r$ with one of the two remaining $LRs$ at random, and return to step \ref{step:train}
\end{enumerate}

\begin{figure}[b]
\begin{center}
\includegraphics[width=1.0\linewidth]{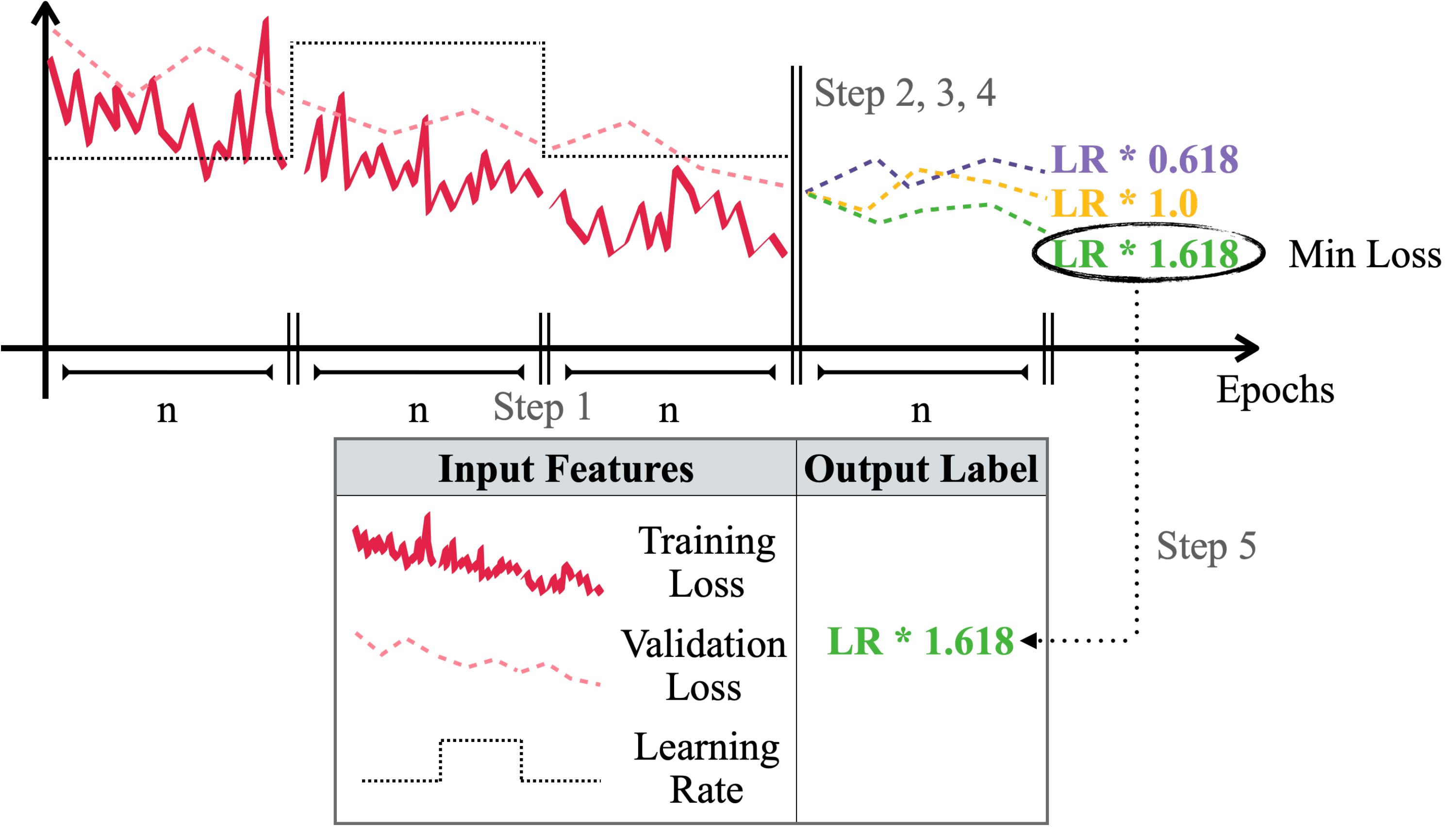}
\end{center}
   \caption{Generating and labeling a data point. In this hypothetical example, increasing the learning rate was found to result in the lowest validation loss after a further $n$ epochs of training. Therefore, the prior $3n$ epochs of training loss, validation loss, and learning rate are assigned a corresponding ground truth label of `increase.'}
\label{fig:data}
\end{figure}

By executing steps \ref{step:train} - \ref{step:truth} we can create one input/ground truth pair. The input features are the training loss, validation loss, and LR during step \ref{step:train}, concatenated with those same features from the previous $2n$ epochs of training. The label is the LR noted in step \ref{step:truth}. This process is depicted in Figure \ref{fig:data}. Steps \ref{step:train} - \ref{step:repeat} continue until training finishes. Assuming the total number of training epochs is $N$, then we get $N/n$ data points from each training procedure. 

We adopt this specific workflow based on the following intuitions: step \ref{step:train}, by providing an identical starting point, ensures a fair comparison between different subsequent LR decisions. Steps \ref{step:increase} - \ref{step:decrease} evaluate the impact of different LR decisions in actual training. We randomly choose different values for $n$ across different data generation runs to avoid bias towards a particular time horizon / dataset size. For step \ref{step:truth} we prefer decisions which lead to lower validation losses for consistency with our learning objective. The random selection process in step \ref{step:repeat} is used to explore a larger search space without causing the loss to diverge. While there could be many alterations to this overall workflow, this form was chosen to generate the data required by our specific problem statement due to its relative simplicity. 

To help ensure generalization, we gathered 12 different computer vision and language tasks - each having a different configuration (dataset, architecture, initial LR ($LR_0$), etc.) as shown in Table \ref{tbl:tasks}. Note that these tasks capture a large variety of different losses, from sparse and per-pixel CE to hinge, adversarial, dice, and multi-task losses. Each of the 12 tasks were trained an average of 42 times. Each training randomly selected an optimizer from \{Adam, SGD, RMSprop\cite{rmsprop}\}, an $LR_0$ $r\in [\text{Min Init LR}, \text{Max Init LR}]$, a value of $n\in [1, 10]$, and then trained for a total of $10n$ epochs. Thus approximately 5050 sample points were collected in total.

\begin{table*}
\begin{center}
\resizebox{\textwidth}{!}{\begin{tabular}{|c|c|c|c|c|c|}
\hline
Task & Task Description & Dataset & Architecture & Max $LR_0$ & Min $LR_0$ \\
\hline\hline
1 & Image Classification & SVHN Cropped \cite{svhn} & VGG19 \cite{vgg} + BatchNorm \cite{batchnorm} & 1e-3 & 1e-5\\
2 & Image Classification & SVHN Cropped & VGG16 \cite{vgg} + ECC \cite{ecc} & 1e-4 & 1e-5\\
3 & Adversarial Training \cite{fgsm} & SVHN Cropped & VGG19 & 1e-3 & 1e-5\\
4 & Image Classification & Food101 \cite{food101} & Densenet121 \cite{densenet} & 1e-2 & 1e-5\\
5 & Image Classification & Food101 & InceptionV3 \cite{inceptionv3} & 1e-2 & 1e-5\\
6 & Multi-Task \cite{multitask} & CUB200 \cite{cub200} & ResNet50 \cite{resnet50} + UNet \cite{unet} & 1e-4 & 1e-5\\
7 & Text Classification & IMDB \cite{imdb} & LSTM & 1e-3 & 1e-5\\
8 & Named Entity Recognition & MIT Movie Corpus \cite{mitmovie} & BERT \cite{bert} & 1e-4 & 1e-5\\
9 & One Shot Learning & omniglot \cite{omniglot} & Siamese Network \cite{siamese} & 1e-3 & 1e-5\\
10 & Text Generation & Shakespear \cite{shakespear}& GRU \cite{gru} & 1e-3 & 1e-5\\
11 & Semantic Segmentation & montgomery\cite{montgomery}& UNet& 1e-4 & 1e-5\\
12 & Semantic Segmentation & CUB200 & UNet & 1e-3 & 1e-5\\
\hline
\end{tabular}}
\end{center}
\caption{Training dataset task overview. Training tasks spanned a variety of domains, datasets, architectures, and learning rates in order to provide ARC with a wide sampling of experiences to learn from.}
\label{tbl:tasks}
\end{table*}

\subsection{Correcting Ground Truth}
\label{subsec:corrections}

Suppose that during step \ref{step:truth} of the data generation process you find that $l_+$, $l_1$, and $l_-$ are 0.113, 0.112, and 0.111 respectively. Due to challenge \ref{challenges:random} it may not be appropriate to confidently claim that decreasing LR is the best course of action.

Luckily, there is one more datapoint we can use to reduce uncertainty. Suppose that during step \ref{step:repeat} we choose to decrease the LR. Then the subsequent step \ref{step:train} is repeating exactly the prior step \ref{step:decrease}. Let $l_-^*$ be the validation loss at the end of step \ref{step:train}. If the relative order of $l_+$, $l_1$, and $l_-$ is the same as the relative order of $l_+$, $l_1$, and $l_-^*$, then we consider our ground truth labeling to be correct (for example, if $l_-^*=0.109)$. On the other hand, if the relative ordering is different (for example, if $l_-^*=0.115$), then random noise is playing a greater role than the LR in determining performance. In that case we take a conservative approach and modify the ground truth label to be `constant LR'.

\subsection{Building the Model}
\label{subsec:model}

For preprocessing, considering challenge \ref{challenges:scale}, we apply z-score normalization to the training and validation losses, and then perform nearest-neighbor resizing to length 100, 200, or 300 (depending on whether we have $n$, $2n$, or $3n$ epochs of history). If only $n$ or $2n$ epochs of prior data were available (very early during training), we zero-prepend the loss vectors to length 300. For LR we normalize by dividing by the first available value, then perform resizing followed, if necessary, by prepending ones to ensure a final length of 300.

\begin{figure*}[t]
\begin{center}
\includegraphics[width=0.9\linewidth]{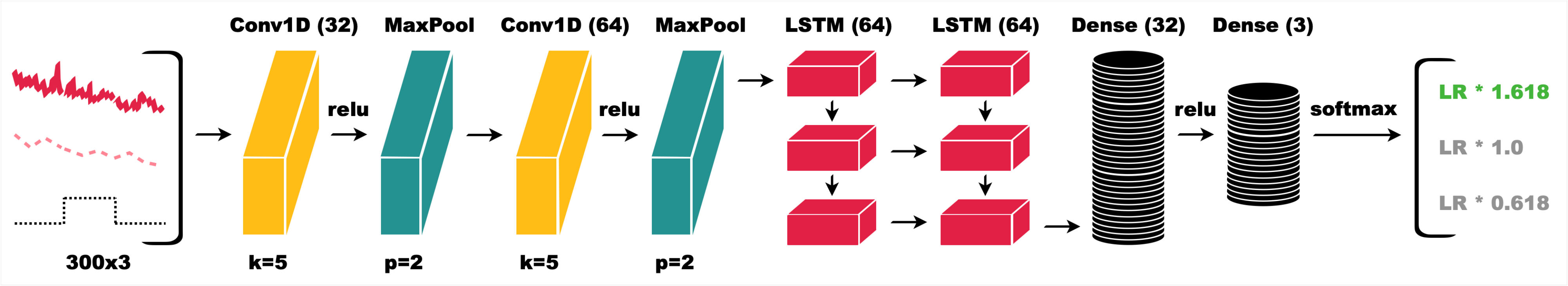}
\end{center}
   \caption{Network architecture used in ARC. The model receives historical measurements of training loss, validation loss, and learning rate. After parsing this information through 8 lightweight layers it produces one of three classes as output (increase, keep, or decrease the current LR).}
\label{fig:network}
\end{figure*}

The ARC model architecture is shown in Figure \ref{fig:network}. It consists of three components: a feature extractor, an LSTM \cite{lstm}, and a dense classifier. The feature extractor consists of two 1D convolution layers, the LSTM of two stacked memory sequences, and the classifier of two densely connected layers. Considering constraint \ref{challenges:footprint}, we chose layer parameters such that the total number of trainable model parameters is less than 80k. Compared to the millions of parameters which are common in current state-of-the-art models, this architecture should add relatively minimal overhead.

To train ARC we used a hybrid loss which averaged classical categorical cross-entropy and a specialized binary cross-entropy loss. For the binary cross-entropy loss we performed a one-vs-all binarization of the task focusing on the `increase LR’ class, since an over-aggressive increase can lead to divergence / NaN values.

The model was trained with the corrected dataset from Section \ref{subsec:corrections}. We leveraged an Adam optimizer with the following parameters: $LR=1e-4$, $\beta_1=0.9$, and $\beta_2=0.999$. Training proceeded with a batch size of 128 for 300 epochs. Once trained, the ARC model can be used to periodically adjust the LR for other models, as we demonstrate in Section \ref{sec:experiments}.

\subsection{Finding the Best ARC Model}

We found that our training procedure produces ARC models with noticeably different behaviors from one another. Ideally we would select the best model based on some metric, however we found that neither loss, accuracy, weighted accuracy, calibration error \cite{calibration}, nor MCC \cite{mcc} on held-out training data were strongly correlated with how well the model would perform on downstream tasks (results in Appendix \ref{secA1}). We believe this indicates that certain LR decisions are much more important than others (for example, perhaps early decision matter more than later ones). This is not captured in aggregating metrics such as accuracy. To circumvent this issue we used real training performance measurements on a separate problem to score the models.

We generated 10 different candidate models by repeating our Section \ref{subsec:model} procedure 10 times. We then used each candidate to train an auxillary task on multiple $LR_0$s, selecting the model leading to the best average task performance across $LR_0$s for further use. In particular, we trained a wide residual network \cite{wrn} on the SVHN Cropped dataset (Adam optimizer) over three different $LR_0$s: 1e-1, 1e-3, and 1e-5. Each candidate was used to perform 5 independent trainings at each $LR_0$. The proxy problem scores for each candidate network can be found in Appendix \ref{secA2}. Note that after an ARC model has been created in this way it can then be used on a variety of different downstream tasks without any further tuning, as we demonstrate in Section \ref{sec:experiments}.

\section{Experiments}
\label{sec:experiments}

In this section, we test how well ARC can guide training tasks on previously unseen datasets and architectures. Specifically, we deploy ARC on two computer vision tasks and one NLP task. For each task, we compare ARC against 4 standard LR schedules: Baseline LR (BLR) - in which LR is held constant, one-cycle Cosine Decay (CD), Cyclic Cosine Decay (CCD), and Exponential Decay (ED). We also compare against 2 more sophisticated LR approaches: Superconvergence (SC) and SGD+Armijo Stochastic Line Search (SLS). For each task we use the same ARC model, invoked once every 3 epochs (which we found to be a good general rule, see Appendix \ref{secA5}).

When confronted with a new dataset or network architecture, it is unclear \textit{a priori} what an ideal $LR_0$ will be. Ideally an LR schedule would therefore be robust against a variety of different $LR_0$s. In order to gain a holistic view of the effectiveness of different schedulers, we use 3 different $LR_0$s for each task. We selected our maximum $LR_0$ such that larger values risked BLR divergence,  and our minimum $LR_0$ such that smaller values led BLR to converge too slowly to be useful. Each training configuration is run 5 times, with median test metric performance (\emph{e.g.} median test accuracy) being reported.

SC has its own process for determining which $LR_0$ to use, leveraging an `LR range test' ahead of the primary training. We therefore run SC on the $LR_0$ value indicated by the range test for each task (range test results in Appendix \ref{secA3}).

SLS computes its LR at every step of training, and thus does not take $LR_0$ as a parameter. We run it 5 times per task and report the median score in comparison to the other methods for each of their $LR_0$ values. We used the official implementation of the SLS optimizer provided by \cite{vaswani2019painless}.

\subsection{Image Classification on CIFAR10}
\label{subsec:cifar}

For our first experiment we trained a model to perform CIFAR10 image classification. We used the same architecture and preprocessing as proposed in \cite{fastcifar}. We trained for 30 epochs (rather than 24 in the original implementation) using an Adam optimizer and a batch size of 128. Three different $LR_0$s were used: 1e-2, 1e-3, and 1e-4. For each $LR_0$, we compare the performance of BLR with the performance of ARC (invoked every 3 epochs), as well as CD, CCD (using the settings proposed in \cite{enas} for CIFAR10), ED ($\gamma=0.9$), SC ($LR_{max}=0.198$ per Appendix \ref{secA3}), and SLS.

The results for all experiment runs are summarized in Table \ref{tbl:cifar10}. The median-run graphs of LR and validation accuracy over time for each method are given in Figure \ref{fig:cifar}. A summary of all runs can be found in Appendix \ref{secA4}.

When the $LR_0$ is sufficiently large (1e-2 and 1e-3), all decaying LR schedulers outperform the baseline LR. From Figure \ref{fig:cifar} (a) and (b), we can see that ARC also decided to decrease the LR. Amongst all LR schedulers, ARC performed third best with the large $LR_0$ (1e-2), and was the best performer with the medium $LR_0$ (1e-3). Interestingly, even when ARC was outperformed by CD, it very closely emulated the CD decay pattern as shown in Figure \ref{fig:cifar} (a).

\begin{figure*}[b]
\begin{center}
\includegraphics[width=\linewidth]{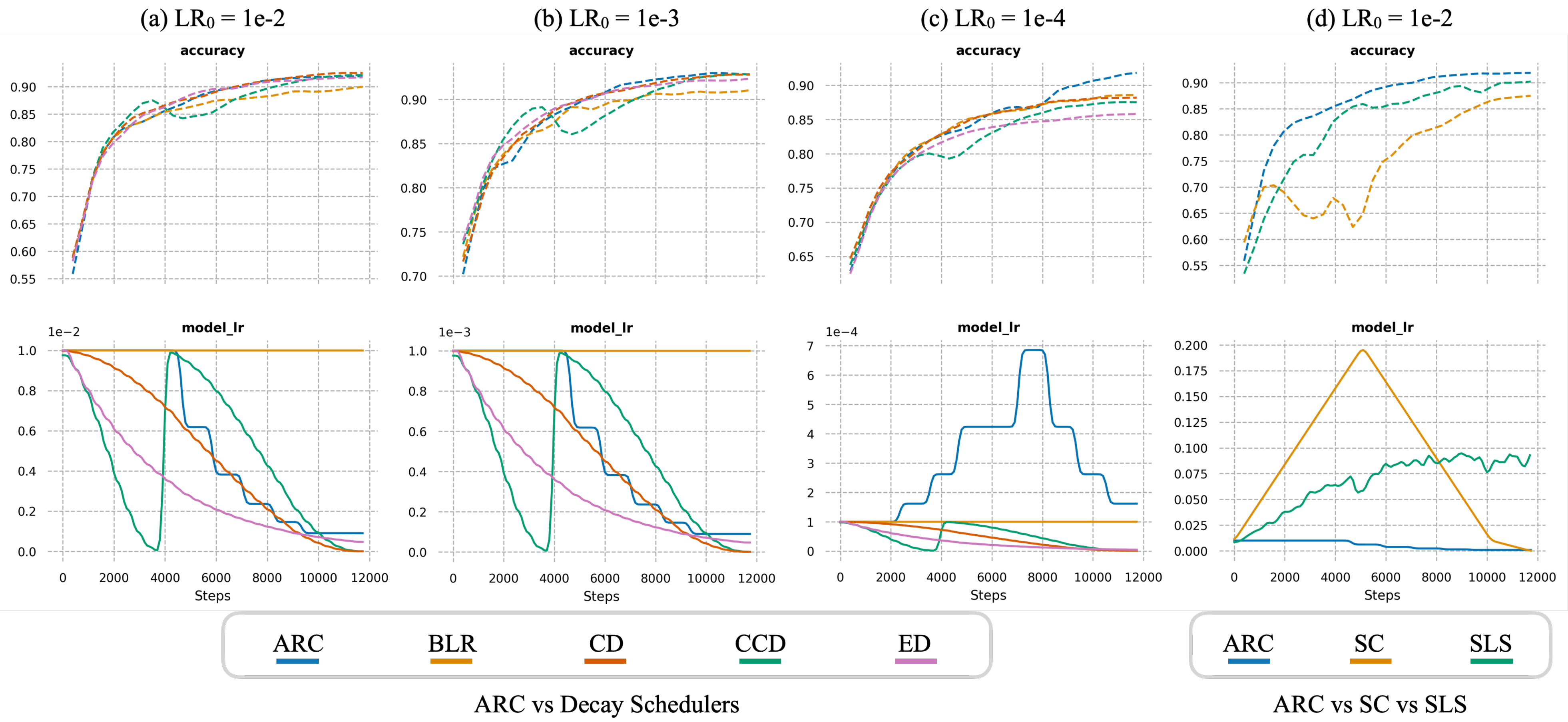}
\end{center}
   \caption{Median run performance (validation accuracy and LR vs training steps) on CIFAR10. For better LR visualization, SC and SLS are separated from the decay schedulers.}
\label{fig:cifar}
\end{figure*}

\begin{table}[b]
\begin{center}
\begin{tabular}{|c|c|c|c|}
\cline{2-4}
\multicolumn{1}{c|}{} & $LR_0$ = 0.01 & $LR_0$ = 0.001 & $LR_0$ = 0.0001\\
\hline
BLR & 90.19 & 91.42 & 88.82\\
CD & \textbf{92.60}* & 92.89 & 88.22\\
CCD & 92.16& 92.90& 87.61\\
ED & 91.73& 92.52& 85.91\\
SC & 87.64 & -- & -- \\
SLS & 90.90 & 90.90 & 90.90 \\
ARC & 92.00& \textbf{93.09}*& \textbf{91.87}*\\
\hline
\end{tabular}
\end{center}
\caption{CIFAR10 test accuracy. Median over 5 runs, with best (highest) values in bold. Note that SLS is duplicated across columns since it is independent of $LR_0$. Statistically significant ($p < 0.05$) improvements over runner-up values are indicated by an *.}
\label{tbl:cifar10}
\end{table}

When the $LR_0$ is small (1e-4), however, the drawback of statically decaying LR schedulers becomes evident: decaying an already small LR damages convergence. In this case CD, CCD, and ED are all beaten by the baseline LR. On the other hand, as shown in Figure \ref{fig:cifar} (c), ARC is able to sense that the LR is too small and increase it, achieving the best final accuracy. Furthermore, even given this difficult $LR_0$, ARC outperforms every BLR configuration (1e-2, 1e-3, and 1e-4). Given that the baseline learning rate in this case represents a raw Adam optimizer, this result demonstrates that ARC can be combined with adaptive optimizers to further improve their performance.

ARC also improves over SC and SLS for this problem, both of which increase LR far too aggressively, as can be seen in Figure \ref{fig:cifar} (d).

\subsection{Object Detection on MSCOCO}
\label{subsec:mscoco}

Our second and most time-consuming task is object detection using the MSCOCO dataset. We downscale the longest side of each image to 256 pixels in order to complete the trainings within a more reasonable computational budget. The RetinaNet \cite{retinanet} architecture was selected for this task. We used a batch size of 32 and trained for a total of 45000 steps, with validation every 1500 steps. We used a momentum optimizer with 0.9 for its momentum value, but kept all other parameters consistent with the official implementation. The configuration for our LR schedulers is the same as in Section \ref{subsec:cifar}, but with $LR_0$ of 0.01, 0.005, and 0.001. The SC $LR_{max}$ was set to 0.01 (see Appendix \ref{secA3}). For this task we use mean average precision (mAP) to benchmark model performance. Note that this task uses localization and focal losses which were not present in the ARC training dataset.

The results for all experiment runs are summarized in Table \ref{tbl:mscoco}. Median-run graphs of LR and mAP over time for each method are given in Figure \ref{fig:mscoco}. A summary of all runs can be found in Appendix \ref{secA4}.

\begin{figure*}[b]
\begin{center}
\includegraphics[width=\linewidth]{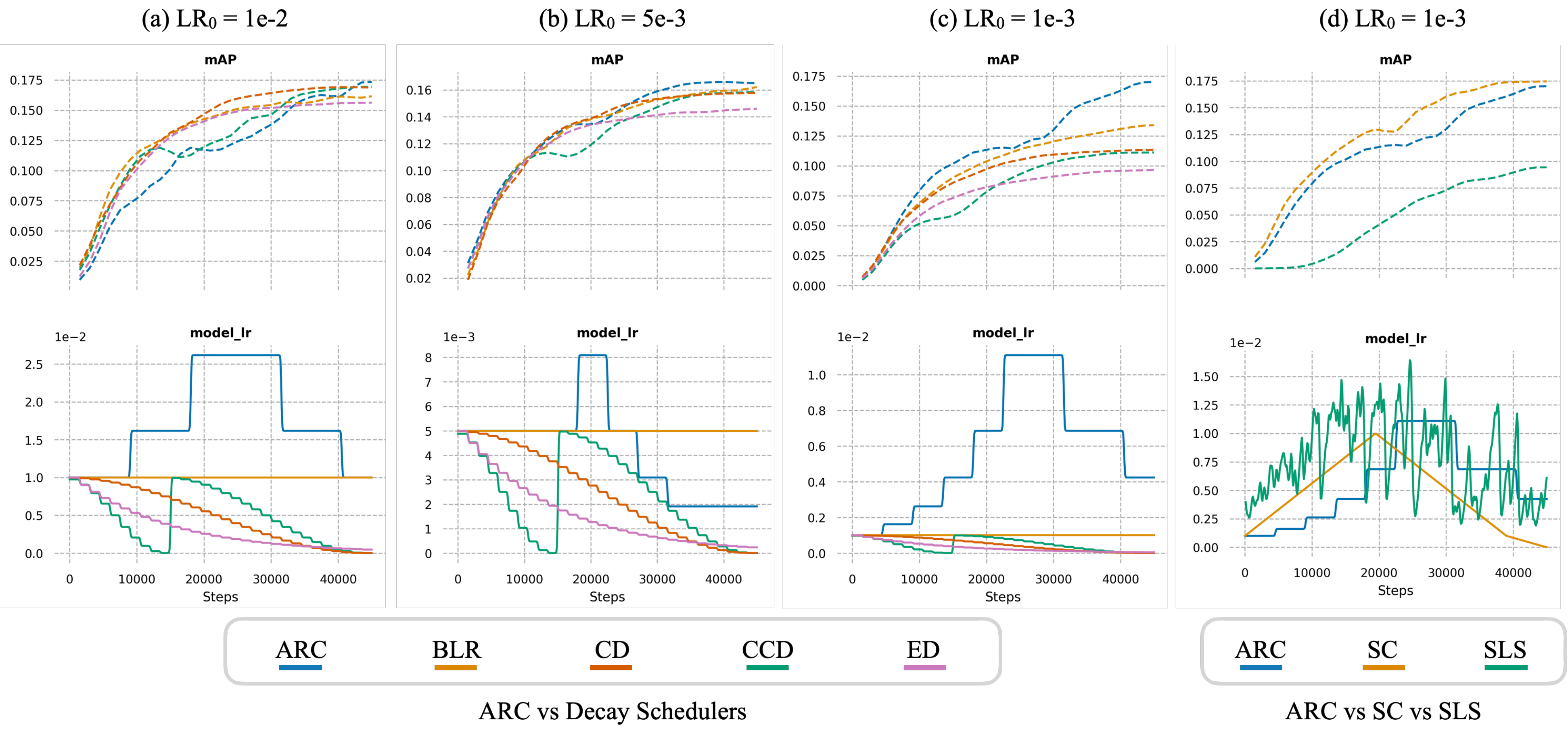}
\end{center}
   \caption{Median run performance (validation mAP and LR vs training steps) on MSCOCO. For better LR visualization, SC and SLS are separated from the decay schedulers.}
\label{fig:mscoco}
\end{figure*}

\begin{table}[t]
\begin{center}
\begin{tabular}{|c|c|c|c|}
\cline{2-4}
\multicolumn{1}{c|}{} & $LR_0$ = 0.01 & $LR_0$ = 0.005 & $LR_0$ = 0.001\\
\hline
BLR & 0.1634 & 0.1629 & 0.1343\\
CD & 0.1692 & 0.1580 & 0.1135\\
CCD & 0.1694 & 0.1584 & 0.1112\\
ED & 0.1563 & 0.1462 & 0.0967\\
SC & -- & -- & \textbf{0.1747}* \\
SLS & 0.0955 & 0.0955 & 0.0955 \\
ARC & \textbf{0.1757} & \textbf{0.1668} & 0.1709\\
\hline
\end{tabular}
\end{center}
\caption{MSCOCO test mAP. Median over 5 runs, with best (highest) values in bold. Note that SLS is duplicated across columns since it is independent of $LR_0$. Statistically significant ($p < 0.05$) improvements over runner-up values are indicated by an *.}
\label{tbl:mscoco}
\end{table}

Interestingly, the largest $LR_0$ (1e-2) we used was not large enough to allow ED to outperform the baseline LR. Unfortunately, larger $LR_0$s were found to lead to training divergence. This exposes a critical limitation of exponential LR decay: the rate of decay needs to be carefully tuned, otherwise the LR can be either too large early on or too small later in training. On the other hand, CD, CCD, and ARC outperform the baseline LR, with ARC achieving the best mAP. Unfortunately SLS does very poorly on this problem despite decent performance on CIFAR.

For the other two smaller LRs (5e-3 and 1e-3), all of the decaying schedules are worse than the baseline LR because they have no mechanism to raise the LR when doing so would be useful. In contrast, ARC can notice this deficiency and increase the LR accordingly - allowing it to achieve strong mAP across the board. SC is able to outperform ARC on this problem, but the two methods are much closer to one another than they are to any runner-up candidates. In fact, from Figure \ref{fig:mscoco} (d) it appears that ARC has learned to naturally mimic SC when it sees it beneficial, although it is limited by the fact that it only executes once every 3 epochs.

\subsection{Language Modeling on PTB}

For our final experiment we move beyond computer vision to verify whether ARC can be useful in natural language processing tasks as well. We performed language modeling using the PTB dataset \cite{ptb} with a vocabulary size of 10000. Our network for this problem leveraged 600 LSTM units with 300 embedding dimensions, and a 50\% dropout applied before the final prediction. Training progressed for 98 epochs, with a batch size of 128 and a sequence length of 20. A Stochastic Gradient Descent (SGD) optimizer was selected, with $LR_0$ values of 1.0, 0.1, and 0.01. Our CCD scheduler used $T_0=14$ and $T_{multi}=2$ such that we could fit 3 LR cycles into the training window. The ED scheduler $\gamma$ value was set to 0.96, and the SC $LR_{max}$ was 23.2 (see Appendix \ref{secA3}). For this task we used perplexity to measure model performance (lower is better).

\begin{figure*}[t]
   \begin{center}
   \includegraphics[width=\linewidth]{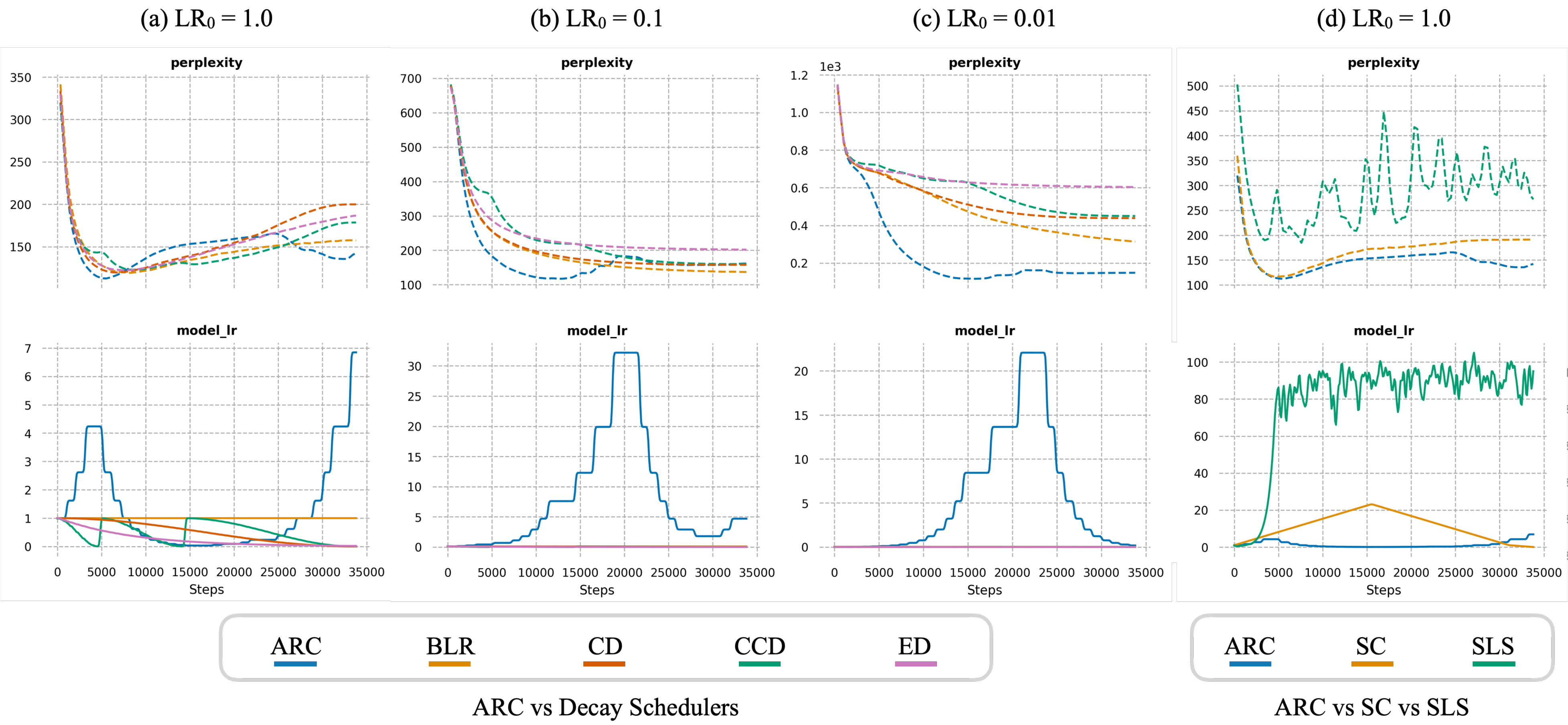}
   \end{center}
      \caption{Median run performance (validation perplexity and LR vs training steps) on PTB. For better LR visualization, SC and SLS are separated from the decay schedulers.}
   \label{fig:ptb}
\end{figure*}

\begin{table}[t]
\begin{center}
\begin{tabular}{|c|c|c|c|}
\cline{2-4}
\multicolumn{1}{c|}{} & $LR_0$ = 1.0 & $LR_0$ = 0.1 & $LR_0$ = 0.01\\
\hline
BLR & 118.3 & 136.7 & 313.5\\
CD & 119.2 & 157.8 & 438.8\\
CCD & 122.6 & 160.0 & 450.1\\
ED & 122.1 & 202.0 & 603.5\\
SC & 116.1 & -- & -- \\
SLS & 150.1 & 150.1 & 150.1 \\
ARC & \textbf{111.3}* & \textbf{116.1}* & \textbf{115.9}*\\
\hline
\end{tabular}
\end{center}
\caption{PTB test perplexity. Median over 5 runs, with best (lowest) values in bold. Note that SLS is duplicated across columns since it is independent of $LR_0$. Statistically significant ($p < 0.05$) improvements over runner-up values are indicated by an *.}
\label{tbl:ptb}
\end{table}

The results for all experiment runs are summarized in Table \ref{tbl:ptb}. Median-run graphs of LR and perplexity over time for each method are given in Figure \ref{fig:ptb}. A summary of all runs can be found in Appendix \ref{secA4}.

ARC had very strong performance on this problem, achieving the best scores regardless of which $LR_0$ it was given. SC also performed well, and SLS once again struggled, though not to the same extent as for object detection.

\section{Open Problems and Opportunities}
\label{sec:limits}

As Section \ref{sec:experiments} demonstrates, ARC can be successfully deployed over a range of tasks, architectures, optimizers, dataset and batch sizes, and $LR_0$s. It does, however, have some open problems along with opportunities for future work which bear mentioning.

One open problem with ARC is that it assumes a constant optimization objective. While this is often the case for real-world problem-solving tasks, it is not true of generative adversarial networks (GANs), where the loss of the generator is based on the performance of an ever-evolving discriminator. Thus ARC, while applicable to many problems, may not be appropriate for all genres of deep learning research. On the other hand, the design of ARC does not technically preclude its use in such cases. We see investigating this as an opportunity in the future.

Another open problem with the current ARC implementation is that it sometimes provides unreliable decisions if queried too frequently. We found that once every 3 epochs is the best frequency (see Appendix \ref{secA5}). The need for 3 epochs may be attributable to the way in which we train ARC using short-, mid-, and long-term data, or else perhaps a consequence of short-horizon bias \cite{wu2018understanding}. Notably, in the only instance where ARC performed worse than SC, it was essentially emulating SC but was unable to do so quickly enough. See Figure \ref{fig:mscoco} (d). It may be possible to avoid this limitation by measuring validation loss more frequently, rather than once at the end of each epoch. We see exploring this possibility as an interesting direction for future work.

\begin{figure*}[t]
\begin{center}
\includegraphics[width=0.85\linewidth]{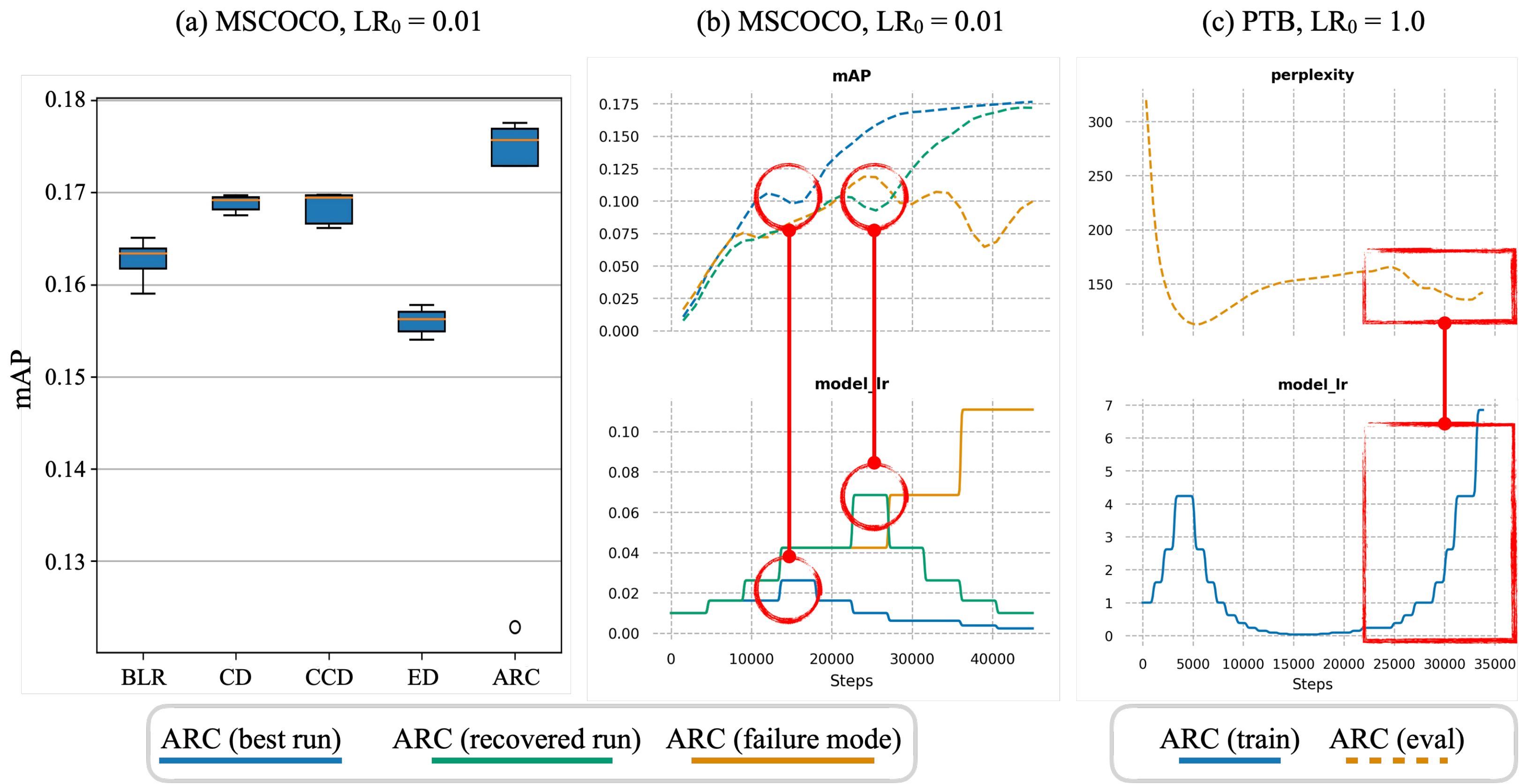}
\end{center}
   \caption{Failure modes and unexpected behaviors. (a) shows the best evaluation performance of 5 independent trainings for different schedulers. (b) shows 3 sample runs from (a) with the outlier run in orange. (c) highlights the median PTB example run from Figure \ref{fig:ptb} (a).}
\label{fig:failures}
\end{figure*}

As for failure modes, just like any other deep learning model, ARC can also make incorrect decisions. For example, when training MSCOCO at an $LR_0$ of 0.01, 4 out of 5 ARC trainings noticeably outperformed all competing schedules as shown in Figure \ref{fig:failures} (a). There was, however, an outlier which performed significantly worse. Its training plot is shown in Figure \ref{fig:failures} (b) alongside with two other ARC training runs. In all 3 of the visualized training runs, ARC raised the LR too aggressively (circled in red in the Figure). In the blue and green runs, ARC automatically detected that the LR was too high and decreased it aggressively, leading to strong final performance. In the orange (failure) case, ARC for some reason doubled down on the excessively large LR, leading to further degradation in performance. Although this is clearly undesirable behavior, it is quite rare - occurring only three times across our total of 45 different experimental configurations/runs (one time in each of the different PTB $LR_0$ configurations). Since real-world applications tend to train multiple models and keep the best one, we do not foresee this being usage-limiting.

Figure \ref{fig:failures} (c) shows an interesting phenomenon which we did not anticipate. After achieving an optimal performance around step 5000, ARC started to drop the LR as might normally be expected to improve performance. However, after step 25000 it changed course and dramatically increased the LR. Comparing ARC's performance with the other schedulers in Figure \ref{fig:ptb} (a), it seems that ARC may be attempting to prevent the model from overfitting. It's not clear whether this is actually a useful strategy, but it's something we hope to investigate more thoroughly in future research.

\section{Conclusion}
In this work we proposed an autonomous learning rate controller that can guide deep learning training to reliably better results. ARC overcomes several challenges in LR scheduling and is complementary to modern adaptive optimizers. We experimentally demonstrated its superiority to popular schedulers across a variety of tasks, optimizers, batch sizes, and network architectures, as well as identifying several areas for future improvement. Not only that, ARC achieves its objectives without tangibly increasing the training budget, adding additional optimization loops, nor introducing complex RL workflows. This is in sharp contrast to prior work in this field, as well as other AutoML paradigms in general. The true test of any automation system is not whether it can outperform any possible hand-crafted custom solution, but rather whether it can provide a high quality output with great efficiency. Given that, the fact that ARC actually does outperform popular scheduling methods while requiring no effort nor extra computation budget on the part of the end user makes it a valuable addition to the AutoML domain.

\backmatter

\section*{Conflict of Interest}

The authors declare that they have no conflict of interest.

\section*{Data Availability Statement}

The datasets used in this study are publicly available for download in their corresponding websites. The source code to reproduce this work has been open-sourced and can be found at \url{https://github.com/fastestimator/ARC/tree/v2.0}.

\begin{appendices}

\section{Common Performance Metrics and ARC}\label{secA1}

Common performance metrics do not accurately predict an ARC model's downstream performance, making it difficult to determine which of several candidate models is the best. We considered 5 different candidate metrics, along with the final models from each run, and the proxy task from Section \ref{sec:proxy}. Our candidate metrics were accuracy, MCC, validation loss, calibration error, and weighted accuracy (specified in Table \ref{tbl:wacc}).

\begin{table}[h!]
\begin{tabular}{cc}
$w_{acc} = \frac{\sum_i{CM[i,i] * \lvert RPM[i,i] \lvert}}{\sum_{i,j}{CM[i,j] * \lvert RPM[i,j] \lvert}}\notag$ &
\begin{tabular}{|c||c|c|c|}
\hline
Predict \textbackslash Actual & Decrease & Constant & Increase \\
\hline\hline
Decrease & $+3$ & $-1$ & $-3$ \\
Constant & $-1$ & $+1$ & $-1$ \\
Increase & $-3$ & $-1$ & $+3$ \\
\hline
\end{tabular}
\end{tabular}
\caption{Weighted Accuracy ($w_{acc}$) formula and corresponding Reward Penalty Matrix. CM is the Confusion Matrix.}
\label{tbl:wacc}
\end{table}

\begin{figure*}[t]
   \begin{center}
   \includegraphics[width=\linewidth]{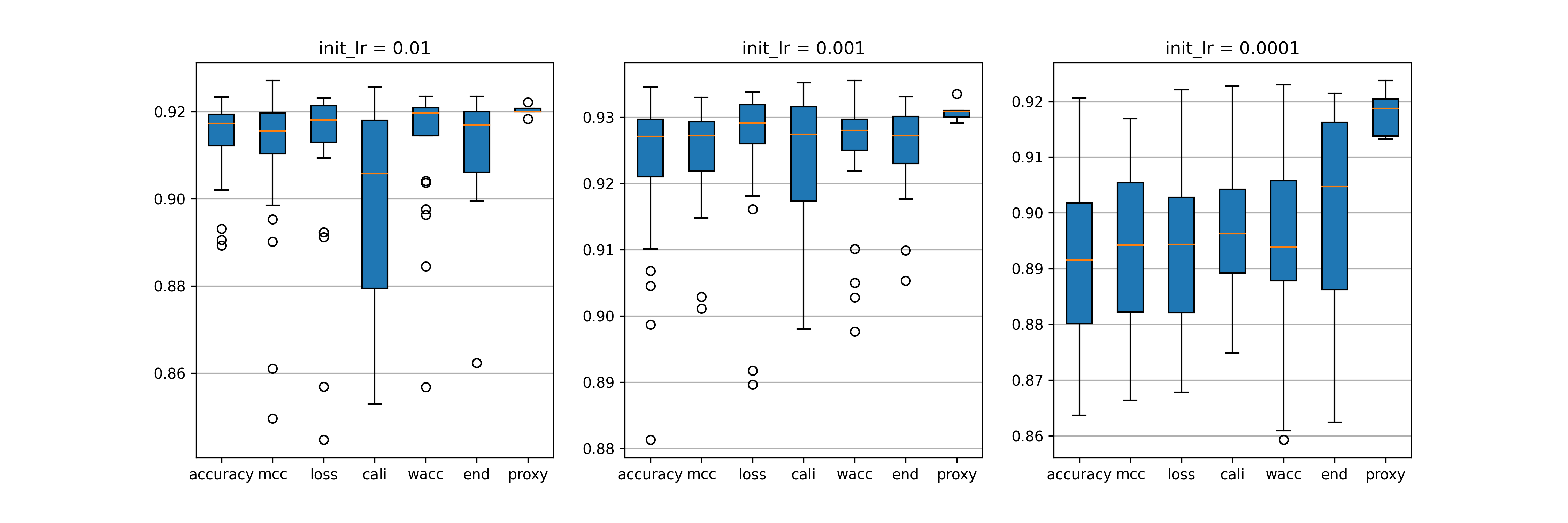}
   \end{center}
      \caption{Evaluating models selected by different metrics. The `proxy' boxplots contain 5 data points each (1 ARC model * 5 training runs). All other boxplots contain 25 data points (5 ARC models * 5 training runs). Y-axis is test accuracy.}
\label{fig:metrics}
\end{figure*}

To test whether any of these metrics were useful, we ran the ARC training procedure a total of 5 times. During those runs, we saved the `best' model according to each of the different metrics, such that in the end we had 5 ARC models for each metric, where models may or may not be the same across metrics (it could be that for one training run, the model with the best accuracy also had the highest MCC). We then used each ARC model to train a 9 layer residual network for 30 epochs on CIFAR10, recording the best evaluation accuracy during training. We tested three different values of $LR_0$: 0.01, 0.001, and 0.0001, and repeated each CIFAR10 training 5 times. The results are summarized in Figure \ref{fig:metrics}. Unfortunately, it seems that none of the simple metrics are strongly indicative of ARC model effectiveness. In fact, simply taking the model from the end of training (epoch 300) was just as effective as any of the metrics we tested. This motivated our use of a proxy problem.

\section{Model Selection via Proxy Task}\label{secA2}
\label{sec:proxy}

\begin{figure*}[t]
   \begin{center}
   \includegraphics[width=\linewidth]{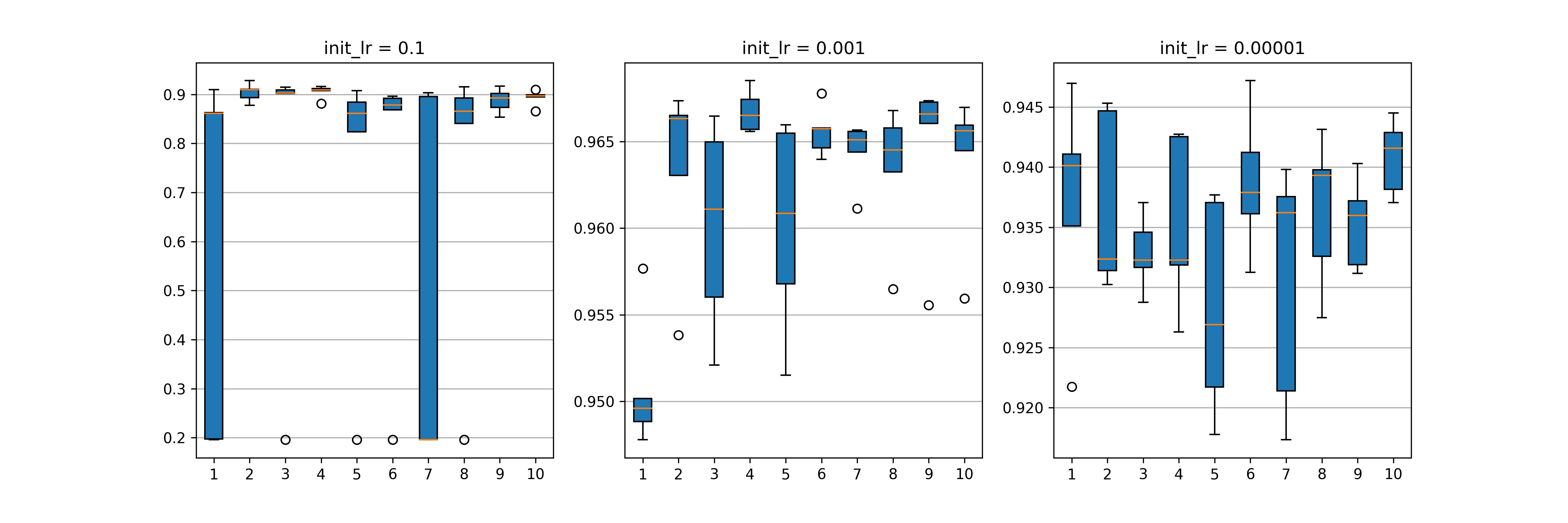}
   \end{center}
      \caption{ARC candidate performance on proxy problem. Each box contains 5 data points (1 ARC model * 5 runs). Y-axis is test accuracy.}
\label{fig:proxy}
\end{figure*}

Since we were unable to find a metric which predicts ARC performance reliably we turn to a proxy task instead. We first train 10 candidate ARC models. We then use each model to train a WideResnet28 architecture on the SVHN Cropped dataset for 30 epochs, repeating each training 5 times. We perform each set of trainings for 3 different values of $LR_0$: 0.1, 0.001, and 0.00001. We gather the best accuracies from each of these SVHN training runs, and select the candidate model with the best mean. Results for each candidate are given in Figure \ref{fig:proxy}. Notice that candidates 1 and 7, which tended to diverge at high LRs, are naturally eliminated by this process. Model 4 was selected for later use.

\section{Superconvergence LR Search}\label{secA3}
\label{sec:range}

\begin{figure*}[h!]
   \begin{center}
   \includegraphics[width=\linewidth]{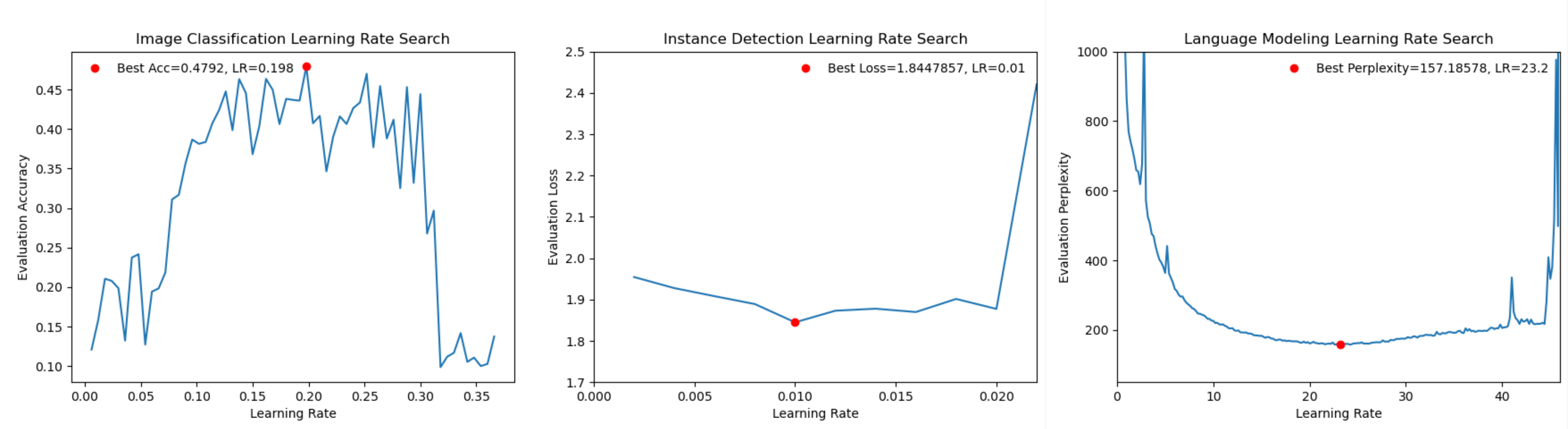}
   \end{center}
      \caption{Superconvergence LR search for various evaluation tasks.}
\label{fig:lrsearch}
\end{figure*}

To use superconvergence LR scheduling, one first needs to run a search routine to determine the ideal maximum LR. For CIFAR10 we use the same LR search routine found in the paper \cite{superconvergence}: LR is increased linearly from 0.0 to 3.0 over 5000 iterations, and evaluation accuracy and LR are plotted accordingly. The max LR is the LR that leads to the maximum accuracy (Figure \ref{fig:lrsearch}). The LR search configuration for other tasks is the same, except that for instance detection LR is increased linearly from 0.0 to 1.0, and for language modeling LR is increased from 0.0 to 100. For instance detection, we use the evaluation loss rather than mAP because the magnitude of the latter is extremely small during early stages of training. As a consequence, we look for the minimum metric score rather than the maximum when searching for the appropriate max LR. We then set $LR_0$ to be $\frac{max LR}{f}$, where $f\in[5,40]$ (consistent with the original paper's experiment section). The $f$ chosen for image classification, instance detection, and language modeling are 19.8, 10.0, and 23.2 respectively, such that their $LR_0$ values match those used in other experiments.

\section{Performance Summaries for all Experiments}\label{secA4}

\begin{figure*}[h!]
   \begin{center}
   \includegraphics[width=0.85\linewidth]{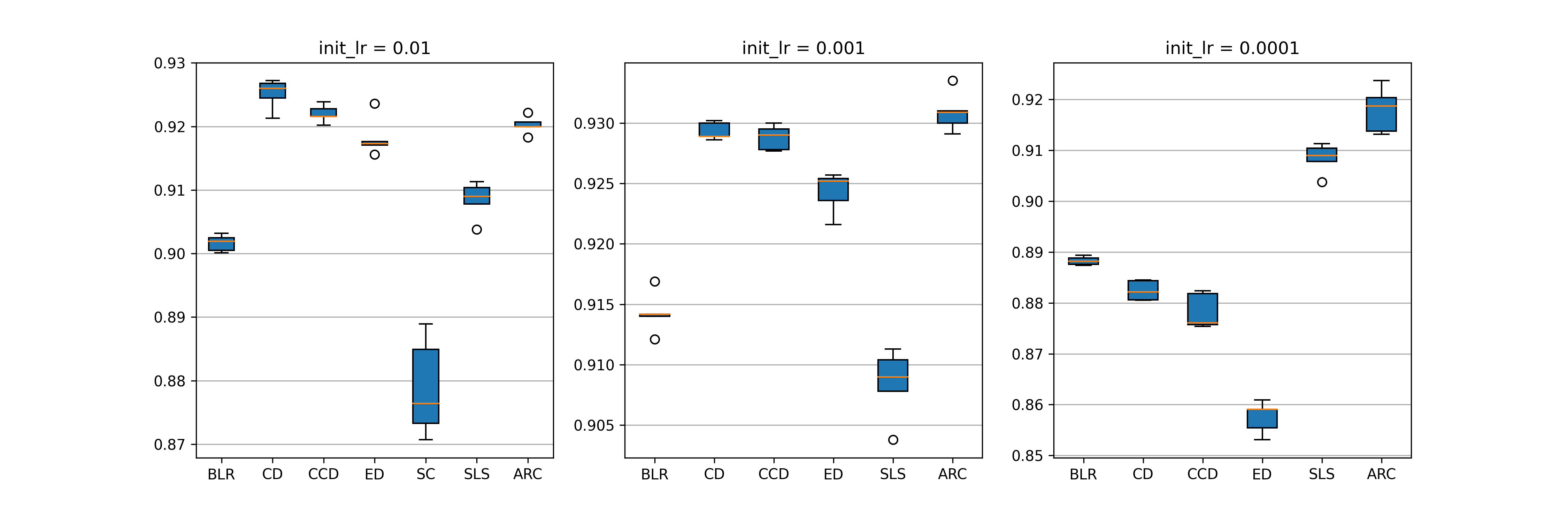}
   \end{center}
      \caption{Image classification summary for all runs. Y-axis is test accuracy (higher is better).}
\label{fig:results_ic}
\end{figure*}

\begin{figure*}[h]
   \begin{center}
   \includegraphics[width=0.85\linewidth]{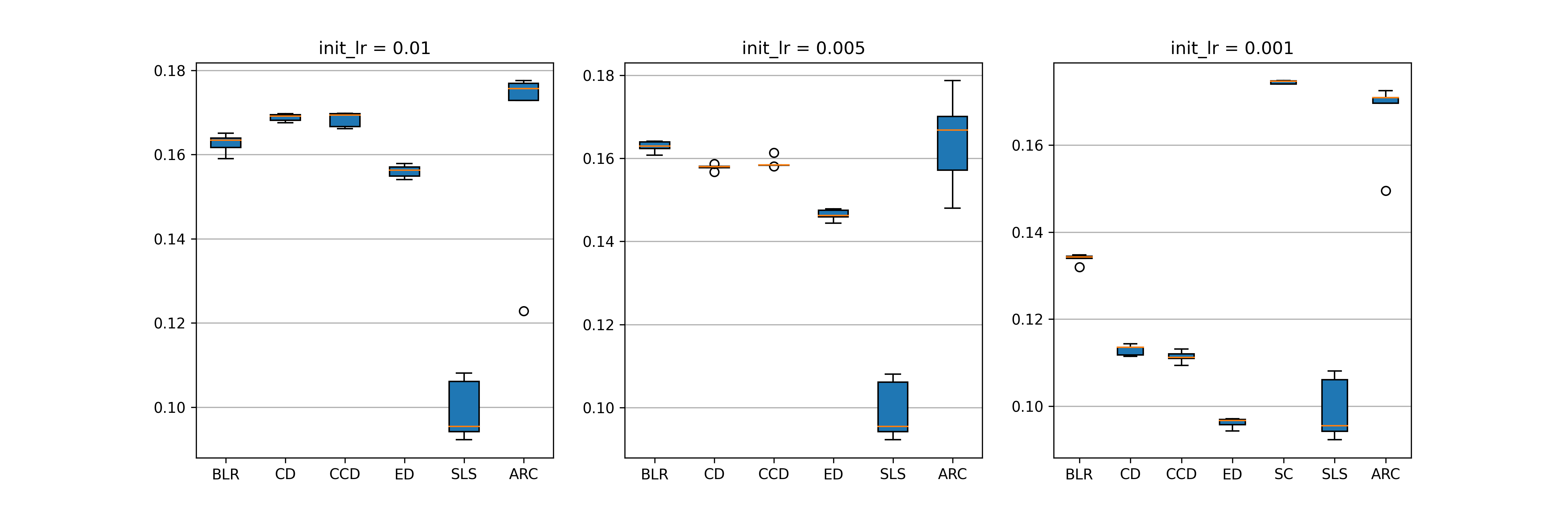}
   \end{center}
      \caption{Instance detection summary for all runs. Y-axis is mAP (higher is better).}
\label{fig:results_id}
\end{figure*}

\begin{figure*}[h]
   \begin{center}
   \includegraphics[width=0.85\linewidth]{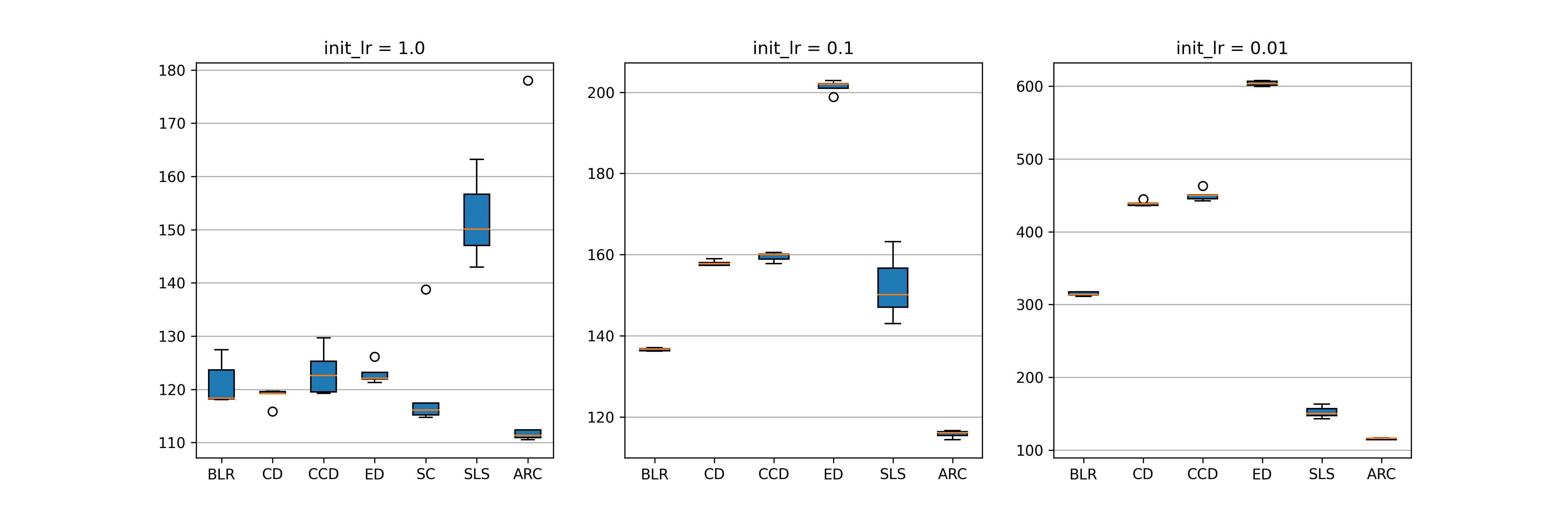}
   \end{center}
      \caption{Language modeling summary for all runs. Y-axis is perplexity (lower is better).}
\label{fig:results_lm}
\end{figure*}

Here we present full performance summaries for all experiments from the paper. Each box in Figures \ref{fig:results_ic}, \ref{fig:results_id}, and \ref{fig:results_lm} contain 5 data points from each of 5 independent runs. SC is only visualized for the $LR_0$ indicated by the LR range test for the particular task (Section \ref{sec:range}).

\begin{figure*}[h]
   \begin{center}
   \includegraphics[width=0.6\linewidth]{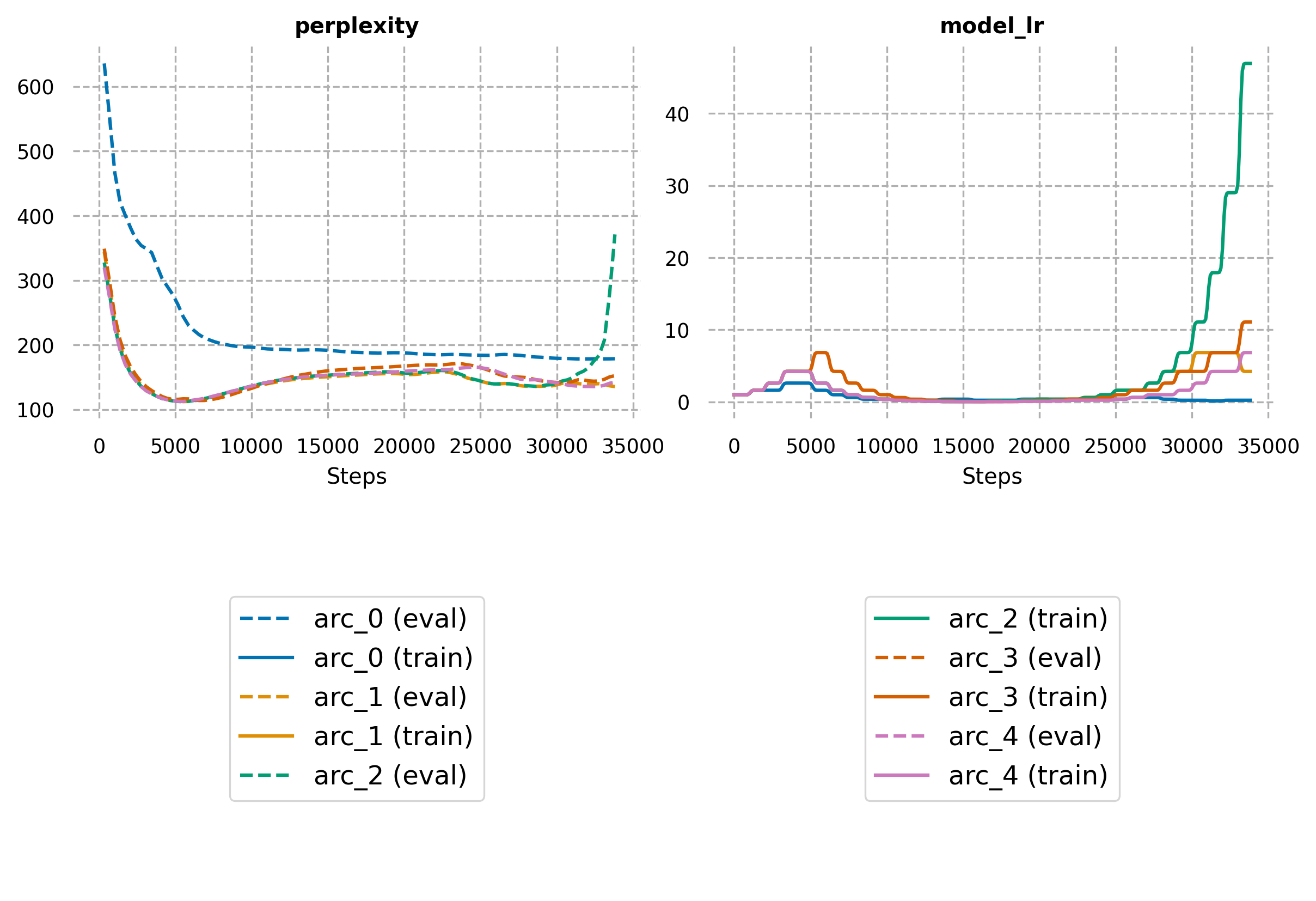}
   \end{center}
      \caption{5 ARC runs for PTB Language Modeling with $LR_0=1.0$. The blue ($\text{arc}_0$) outlier run had a dramatically worse initialization than the other 4 runs, but nonetheless made good learning rate adjustments.}
\label{fig:ptbarc}
\end{figure*}

Note that there is an extreme outlier for ARC in PTB language modeling with $LR_0=1.0$ (Figure \ref{fig:results_lm}). As Figure \ref{fig:ptbarc} demonstrates, this is a result of an extremely unlucky initialization rather than any mistake on the part of ARC.

\section{ARC Invocation Frequency}\label{secA5}

\begin{figure*}[h!]
   \begin{center}
   \includegraphics[width=0.85\linewidth]{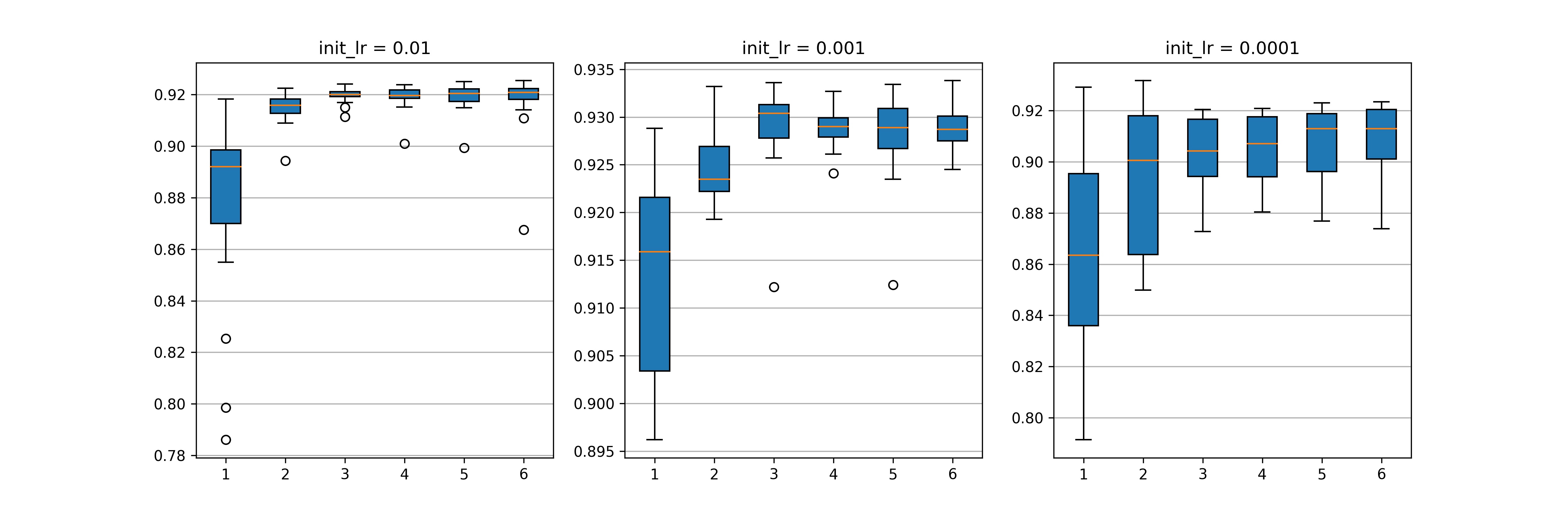}
   \end{center}
      \caption{Effect of ARC frequency on final performance on CIFAR10. Each box contains 25 data points (5 ARC models * 5 runs). Y-axis is test accuracy.}
\label{fig:Frequency}
\end{figure*}

ARC executes at the per-epoch rather than per-step timescale, but this leaves the question of how many epochs should elapse in between invocations. An easy answer would be `as often as possible', but invocations which are too frequent may lead to greater instability or vulnerability to short-horizon bias. To find an ideal frequency we trained 5 ARC models, and then deployed them on top of a 9 layer residual network to train on the CIFAR10 dataset at $LR_0$ values of 0.01, 0.001, and 0.0001. For each $LR_0$ we experimented with invoking ARC once every $n$ epochs for $n\in[1,6]$. Every experimental configuration was repeated 5 times, with maximum accuracies summarized in Figure \ref{fig:Frequency}. For $n\in[1,3]$ the training proceeded for 30 epochs. For $n\in[4,6]$ we trained for $10n$ epochs such that each configuration would have an equal opportunity to adjust the LR. From the results in Figure \ref{fig:Frequency} we conclude that once every 3 epochs is the most frequent invocation schedule we can currently support without harming performance.

\end{appendices}

\bibliography{arc}



\end{document}